\documentclass[letterpaper, 10 pt, conference]{ieeeconf}  

\IEEEoverridecommandlockouts                              

\usepackage{graphicx}
\usepackage{float}
\usepackage{wrapfig}
\usepackage{subfig}

\usepackage{amsthm}
\usepackage{amsmath,amssymb}
\usepackage{mathrsfs}
\usepackage{hyperref}

\usepackage[T1]{fontenc}
\usepackage{aecompl}

\usepackage{cite}
\usepackage{url}

\usepackage{bm}

\usepackage{diagbox}
\usepackage{multirow}
\usepackage{colortbl}
\usepackage{hhline}

\usepackage{xcolor}

\definecolor{orange}{RGB}{255, 250, 205}
\definecolor{royalblue}{RGB}{135, 206, 250}
\definecolor{red}{RGB}{255, 106, 106}
\definecolor{green}{RGB}{152, 251, 152}
\definecolor{brown}{RGB}{205, 186, 150}

\title{\LARGE \bf
Stimulate the Potential of Robots via Competition 
}

\author{Kangyao Huang$^{1}$, Di Guo$^{2}$, Xinyu Zhang$^{3}$, Xiangyang Ji$^{4}$, Huaping Liu$^{1\dag}$
\thanks{This work was supported by the National Natural Science Foundation of China under Grant 62025304.}
\thanks{$^{\dag}$Huaping Liu is the corresponding author.}
\thanks{$^{1}$K. Huang and H. Liu are with the Department of Computer Science and Technology, Tsinghua University, Haidian District, Beijing, 100084, P. R. China,
        {\tt\small huangky22@mails.tsinghua.edu.cn}, {\tt\small hpliu@tsinghua.edu.cn}}%
\thanks{$^{2}$D. Guo is with the School of Artificial Intelligence, Beijing University of Posts and Telecommunications, Beijing, China, {\tt\small guodi.gd@gmail.com}}%
\thanks{$^{3}$X. Zhang is with the School of Vehicle and Mobility, Tsinghua University, Beijing, 100084, P. R. China, {\tt\small xyzhang@tsinghua.edu.cn}}%
\thanks{$^{4}$X. Ji is with the Department of Automation, Tsinghua University, Beijing, 100084, P. R. China, {\tt\small xyji@tsinghua.edu.cn}}%
}

\begin{document}

\maketitle
\thispagestyle{empty}
\pagestyle{empty}

\begin{abstract}

It is common for us to feel pressure in a competition environment, which arises from the desire to obtain success comparing with other individuals or opponents. Although we might get anxious under the pressure, it could also be a drive for us to stimulate our potentials to the best in order to keep up with others. Inspired by this, we propose a competitive learning framework which is able to help individual robot to acquire knowledge from the competition, fully stimulating its dynamics potential in the race. Specifically, the competition information among competitors is introduced as the additional auxiliary signal to learn advantaged actions. We further build a Multiagent-Race environment, and extensive experiments are conducted, demonstrating that robots trained in competitive environments outperform ones that are trained with SoTA algorithms in single robot environment.

\end{abstract}


\section{Introduction}
It has been demonstrated that competition can help improve the physical effort tasks \cite{DiMenichi2015}. For example, multiple race athletes often have the ability to achieve better results in competition that exceed their performance in individual training. Currently, competitive games have been well-studied in multi-agent reinforcement learning (MARL) field, like the professional-level performance of gaming agents implemented in StarCraft \uppercase\expandafter{\romannumeral2} \cite{Vinyals2019}, gFootball\cite{Kurach2020}, and Honor of Kings\cite{Wei2022}. In these studies, researchers pay more attention to the entire team performance, such as the win rate in a mixed-competitive game, converging to Nash Equilibrium (NE) in zero-sum games\cite{Leonardos2021}, or interactions and communication among cooperative agents\cite{Gupta2017,Cheng2024,Cheng2023,Huang2022a,Huang2021a}. However, the potential benefits of leveraging competition information to improve the individual performance is generally overlooked.

\begin{figure}
    \centering
    \includegraphics[width=3.4in]{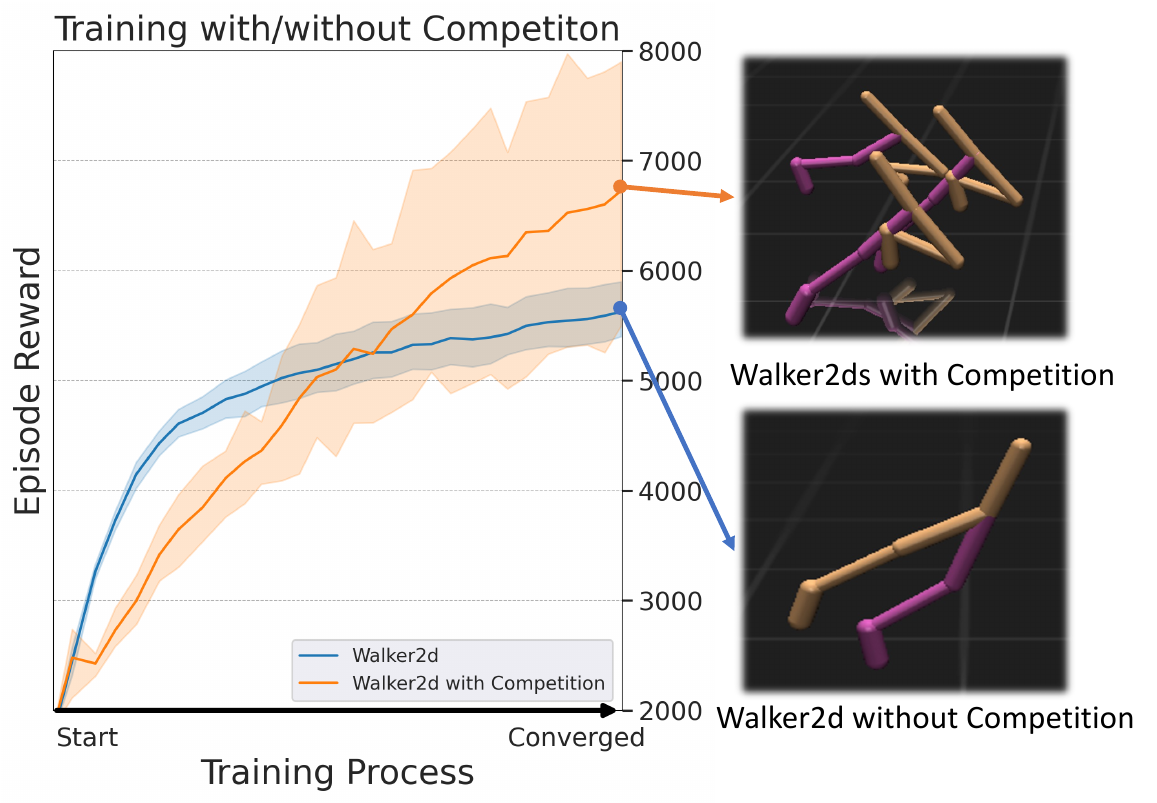}
    \caption{Episode reward comparison between competitive and non-competitive \texttt{Walker2d} environment. The performance of \texttt{Walker2d}s trained with competition can reach 120$\%$ of the baseline.}
    \label{fig:sm-compare}
\end{figure}

In this study, we focus on how to leverage the competition information among multiple robots to facilitate individual robot learning. Our approach aims to understand the connection between favorable actions and rewards. We propose a competitive learning framework which is able to help individual robot to acquire
knowledge from the competition, fully stimulating its dynamics
potential in the race.  Specifically, the competition information among competitors is introduced as the additional auxiliary signal to enhance the learning process. As is shown in Fig.\ref{fig:sm-compare}, we find that under the competitive multi-agent environment, the individual robots can obtain higher rewards and continue to stimulate their dynamics potential compared with non-competitive environment, breaking through the baseline. The main contributions of the work are summarized as follows:

\begin{itemize}
    \item We propose a competitive reinforcement learning framework in which we use additional competition data among multiple robots to enhance the performance of an individual agent.
    \item We suggest that even by incorporating fundamental raw competitive data into the observation as supplementary auxiliary signals, and maintaining the reward mechanism unchanged, the untapped potential of individual robot can be further stimulated. 
    \item We build a set of self-interest competition environments called  \href{https://github.com/KJaebye/Multiagent-Race}{Multiagent-Race} \footnote{https://github.com/KJaebye/Multiagent-Race}. We investigate how varying numbers of competitors and competitive signals influence the learning performance, and a $20\%$ improvement has been gained over SoTA with the proposed framework (Table.\ref{table:compare}).
\end{itemize}

\section{Related Work}
\subsection{Competition in Multi-agent Task}
There are many types of competition tasks in MARL. Zero-sum games can be summarised in a linear programming (LP) formulation to address Nash equilibrium (NE) problem\cite{Yang2020}. General-sum games might contain cooperation or team-level competition that might exist multiple NE points\cite{Gronauer2022}. Many MARL studies are based on particle dynamics simulators\cite{RyanLoweYIWUAvivTamarJeanHarbOpenAIPieterAbbeel2017, Yu2021}, such as MPE (Multi-Agent Particle Environments). Besides, a few are based on well-developed game engines\cite{Vinyals2019, Wei2022}. And some interactive tasks are based on robotics simulators. For example, researchers build a series of competitive and adversarial environments in MuJoCo\cite{Bansal2018}, involving lots of physical confrontation environments, control strategies for physically simulated two-player competitive sports \cite{Won2021}. 


The above mentioned tasks are zero-sum or general-sum games that pay more attention to the entire team performance. However, in a self-interested game, each player strives to maximize their own utility or payoff without considering the overall outcome or cooperation with other players \cite{Blumenkamp2020}. In the proposed work, we aim to facilitate individual robots in a continuous action space. There are a series of specialized algorithms dealing with continuous action spaces \cite{Silver2014,Erez2016,Abbeel2015,Schulman2017,Haarnoja2018,Fujimoto2018a,Zhang2023}, and many benchmarks have been established\cite{Raffin2020, Huang2023, Weng2022, Fujita2021}. In our work, we use Proximal Policy Optimization (PPO)\cite{Schulman2017} and its multi-agent variant to learn a continuous action task.


\subsection{Learning from Competitive and Adversarial Data}

Contrastive learning (CL) has been widely used in word and sentence embedding in NLP\cite{Gao2021}, image classification in CV\cite{Chen2020}, and implicit collaborative filtering in information retrieval (IR). It is able to extract meaningful representations through positive and negative data pairs. Furthermore, generative adversarial network (GAN) also has drawn significant attention in recent years\cite{Creswell2018} for selecting negatives, while the confrontation between the generator and discriminator may not converge to the ideal NE, and there is still potential for further exploration and improvement in the adversarial negative sampling method\cite{Wang2017}. Moreover, some work introduces external disturbance from another adversarial robot to improve the robustness of robotic manipulation tasks\cite{Jian2021}.

Our idea shares fundamental similarities with the approach of adversarial learning and contrastive learning\cite{Xu2022,Le-Khac2020}. We construct the competitive scenario that generates comparative data between opponents and learn features from it.


\section{Problem Formulation}
\label{sec:formulation}
In general, agents can be trained to focus on specific skills by modifying the reward mechanism. In more cases, however, we do not want to change the reward mechanism since performance is somewhat sensitive to the reward, and inappropriate rewards might drown out correct reward signals. It comes to a problem: Can we acquire knowledge from raw comparative information, to surpass the results of normal training?

We first formulate the single-agent continuous action task as a Markov Decision Process (MDP), which can be described as a tuple $\left \langle {S}, {A}, {P}, {R} \right \rangle $. ${S}$ is the state set of a system or environment. ${A}$ is the action set that the agent can take. ${P}$ is the state transition function, and ${P}\left( s'|s, a \right)$ represents the transition probability distribution of the system when transiting into the next state ${s}' \in {S}$ from state $s\in {S}$ after taking an action $a\in {A}$. ${R}$ is the reward function. $r={R}\left(s, a \right)$ is the reward given to agent after taking action $a\in{A}$ under $s\in{S}$. The aim is to find a policy $\pi^{*}$, to maximize the expected value of cumulative discount rewards:
\begin{equation}
    \begin{aligned}
        \pi^{*}&=\mathrm{arg} \underset{\pi}{\mathrm{max}}J_{\pi}\\
        J_{\pi}&=\sum_{t=0}^{\infty}\gamma^{t}\cdot \mathbb{E}\left [ R(s_t, a_t)|s_{0}, \pi \right ]
    \end{aligned}
    \label{eq:mdp}
\end{equation}
where $\gamma$ represents the extent to which the agent discounts future awards.

In this work, we consider the multi-agent scenario. We describe the corresponding MDP as a tuple $\left \langle \mathcal{S}, \mathcal{A}, {P}, {R} \right \rangle $ that includes $N$ number of agents. $\mathcal{S}, \mathcal{A}$ are state and action tuples, respectively, where $\mathcal{S}=(S_{1}, S_{2}, \cdots, S_{N})$, $\mathcal{A}=(A_{1}, A_{2}, \cdots, A_{N})$. $P$ and $R$ remain the same as settings in MDP because agents are totally homogeneous and there is no physical interaction between competitors. In this regard, the objective function in (\ref{eq:mdp}) can be extended as 
\begin{equation}
\hat{J}_{\pi} =\frac{1}{N} \sum_{i=1}^{N} \sum_{t=0}^{\infty}\gamma^{t}\cdot \mathbb{E}\left [ R(s_t^{i}, a_t^{i})|s_{0}^i, \pi \right ] 
\end{equation}
where $s_t^i\in S_i$ and $a_t^i\in A_i$ are the state and action of the $i$-th agent. However, the above setting only facilitates the parallel exploration, but does not provide extra benefits due to the lack of interaction between agents.

In practical training, other agents for the $i$-th agent always can provide information which can be augmented to the state. We use $o_t^i$ to denote the competitive information which can be observed from other agents, and may form a new state as 
\[
\bar{s}_t^i = [s_t^i, o_t^i]
\]
and the objective function becomes
\begin{equation}
\bar{J}_{\bar{\pi}} =\frac{1}{N} \sum_{i=1}^{N} \sum_{t=0}^{\infty}\gamma^{t}\cdot \mathbb{E}\left [ R(\bar{s}_t^{i}, a_t^{i})|\bar{s}_{0}, \bar{\pi} \right ] 
\end{equation}
and the optimization problem is changed as
\[
\bar{\pi}^{*} = \arg \mathop {\max }\limits_{{\bar{\pi}}} \bar{J}_{\bar{\pi}}
\]

What we hope to address is therefore answer if the introduction of the extra state information $o_t^i$ could be beneficial and lead to 
\[
\bar{J}_{\bar{\pi}^*} > J_{\pi^*}
\]

If the results hold, then we may develop a set of new multi-agent competitive learning methods for the single agent. Please note that $o_t^i$ can be raw measurement information or encoding features from observation of other agents.

\section{Methodology}
\label{sec:method}
\begin{figure}
    \includegraphics[width=3.4in]{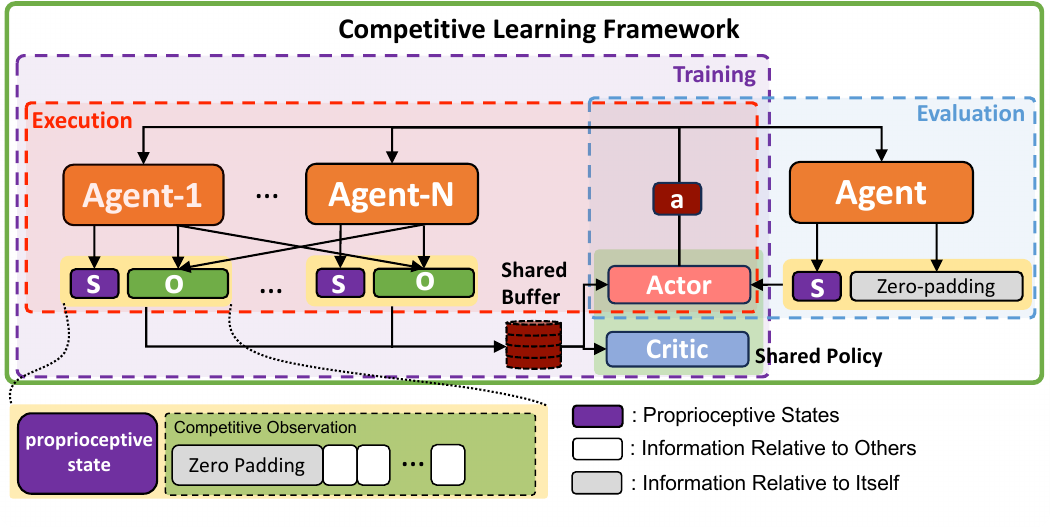}
    \caption{Framework for learning knowledge from comparative information. Where $a$ denotes actions, $s$ represents the proprioceptive state, and $o$ is the competitive observation.}
    \label{fig:framework}
\end{figure}

\subsection{Framework}
We propose a framework to exploit competitive information among multiple agents, shown in Fig.\ref{fig:framework}. The main purpose is to acquire knowledge from comparative data. We run a multi-agent competitive task to obtain additional comparable information and then distinguish positive or negative data for better training. It is akin to the contrasting representation between anchor/positive/negative samples. Using the \texttt{Walker2d} scenario as an illustration, shown in Fig.\ref{fig:walker2d_sim}, if a walker demonstrates exceptional performance, it must be the fastest participant in the race. Concurrently, the robot gathers distinctive relative details, like maintaining a consistently positive relative speed compared to others. This situation creates an auxiliary signal connecting high-speed running and favorable data (comprising state-action-reward pairs).

The framework includes mainly two parts: Training and Evaluation. We train robots with competitive observation during the training stage, while only proprioceptive state is required during the evaluation stage.

\subsection{Policy Training $\&$ Experience Sharing}
The policy training part conducts competitive tasks among multiple agents: agents collect full information involving proprioceptive state and competitive observation. The competitive information vector is a concatenation of relative observation against rivals.

We use the classical actor-critic algorithm PPO and its multi-agent variant\cite{Yu2021} as the algorithmic benchmark because PPO has shown generally favorable results in continuous action control tasks. It becomes normal single-agent learning when there is only one robot in the race.

Since robots are completely homogeneous and exclusively self-interested, we adopt a shared policy among robots. During the training stage, data are sampled distributedly, but the policy is trained centrally since we only maintain one shared policy (actor) $\pi \left(a^{i}\mid \bar{s}^{i};\bm{\theta} \right)$ for all agents, and one shared value (critic) $v(\bar{s}^{i};\bm{\phi})$ to approximate value function $V_{\bm{\phi}}$, where $\bm{\theta}$ and $\bm{\phi}$ denote parameters of the shared policy/value network. Simultaneously, the experience replay buffer $\mathcal{D}$ is also shared across agents to aggregate all experience $\tau=\{\bar{s}, a, r\}$, $r=R(\bar{s}, a)$ is the reward given to agents after taking action $a$ under newly designed state $\bar{s}$. In the training part, critic learns an optimal $\bm{\phi^{*}}$:

\begin{equation}
    \begin{aligned}
        \bm{\phi^{*}}=\mathrm{arg} \underset{\bm{\phi}}{\mathrm{min}} \frac{1}{\left \| \mathcal{D}\right\| T }\sum_{\tau\in \mathcal{D}}^{}\sum_{t=0}^{T} \left ( V_{\bm{\phi}}(\bar{s}_{t}) - R(\bar{s}_t, a_t) \right )^{2}
    \end{aligned}
    \label{eq:critic}
\end{equation}

Here the competitive observation $o$ is introduced as an additional auxiliary signal where we have $\bar{s}=[s, o]$, helping the critic to get a more accurate estimation of experience. We also use GAE(General Advantage Estimation) \cite{Schulman2016} to estimate the advantages of actions $\hat{A}$. Then, we compute the objective function to update the policy:

\begin{equation}
    \begin{aligned}
        L(\bar{s}&, a, \bar{\alpha}, \theta^{new}, \theta^{old}) = \\& \min
        \Bigl( \frac{\pi_{\theta^{new}}(a \mid \bar{s})}{\pi_{\theta^{old}}(a \mid \bar{s})} \hat{A}^{\pi_{\theta^{old}}}(\bar{s}, \bar{\alpha}), \\
        &\mathrm{clip} \Bigl( \frac{\pi_{\theta^{new}}(a \mid \bar{s})}{\pi_{\theta^{old}}(a \mid \bar{s})}, 1 - \epsilon, 1 + \epsilon \Bigr) \hat{A}^{\pi_{\theta^{old}}}(\bar{s}, \bar{\alpha}) \Bigr)
    \end{aligned}
    \label{eq:actor}
\end{equation}

We define the probability ratio between the shared new policy and the old one as $\frac{\pi_{\theta^{new}}(a\mid \bar{s})}{\pi_{\theta^{old}}(a\mid \bar{s})}$. $\bar{\alpha}$ is the action set of all agents, excluding the current agent. Hyperparameter $\epsilon$ is to limit the difference between old and new policies within a small range. 

By sharing experience buffer $\mathcal{D}$, we attain a contrasting effect from competitive data during the critic training. This strengthens the association between rewards and positive/negative actions. The process described by (\ref{eq:critic}) is similar to the step of labeling positive and negative samples for training in CL\cite{Chen2020}. The difference lies in the fact that we utilize reinforcement learning reward measurement that naturally provides label-like signals to label competitive messages. Through incorporating the contrast provided by the multi-agent competition, competitive representation facilitates a clearer and more explicit understanding of the relationship between state and reward. 

\subsection{Robot Observation Construction}
We tailor observation for the robot in order to appropriately introduce the competitive information. Different from the previous study that employs global information\cite{RyanLoweYIWUAvivTamarJeanHarbOpenAIPieterAbbeel2017}, we only consider the partial opponent observation that is more readily obtainable in real-world scenarios. 

As illustrated at the bottom of Fig.\ref{fig:framework}, we consider the new state of the $i$-th agent $\bar{s}^{i}=[s^{i}, o^{i}]$ to be a concatenation of proprioceptive state $s^{i}$ and competitive observation $o^{i}$, or we call it competitive information which is obtained by comparing with other participants. The first part refers to the sensory feedback that robot receives from joints and muscles, providing measurements of motions and torques. The second part encompasses relative information that agents treat as competitive pressure. We denote $o^{i}=[o^{i1}, o^{i2},...,o^{ij},...,o^{iN}]$, where $o^{ij}$ is the observation of the agent $i$ regarding to agent $j$. Competitive observation can be any differentiable signals or any other comparable features. In this work, competitive information of the $i$-th robot is a concatenation of differences in velocities and displacements $o^{ij}=[x^{j}-x^{i}, v^{j}-v^{i}]$, which is related to all robots, including itself. $x^{i}$, $x^{j}$ and $v^{i}$, $v^{j}$ represent displacement and velocity of the $i$-th and $j$-th robot, respectively. Thus, the length of new state $\bar{s}$ depends on the number of participants in one race.

\subsection{Evaluation}
During the evaluation part, competitive messages are unnecessary, and agents exclusively rely on the proprioceptive state. To maintain alignment with the input dimensions of the neural network, we do zero-padding on the missing dimensions.

\section{Emperical Results}
\label{sec:results}

\begin{figure}
    \centering
    \includegraphics[width=3.4in]{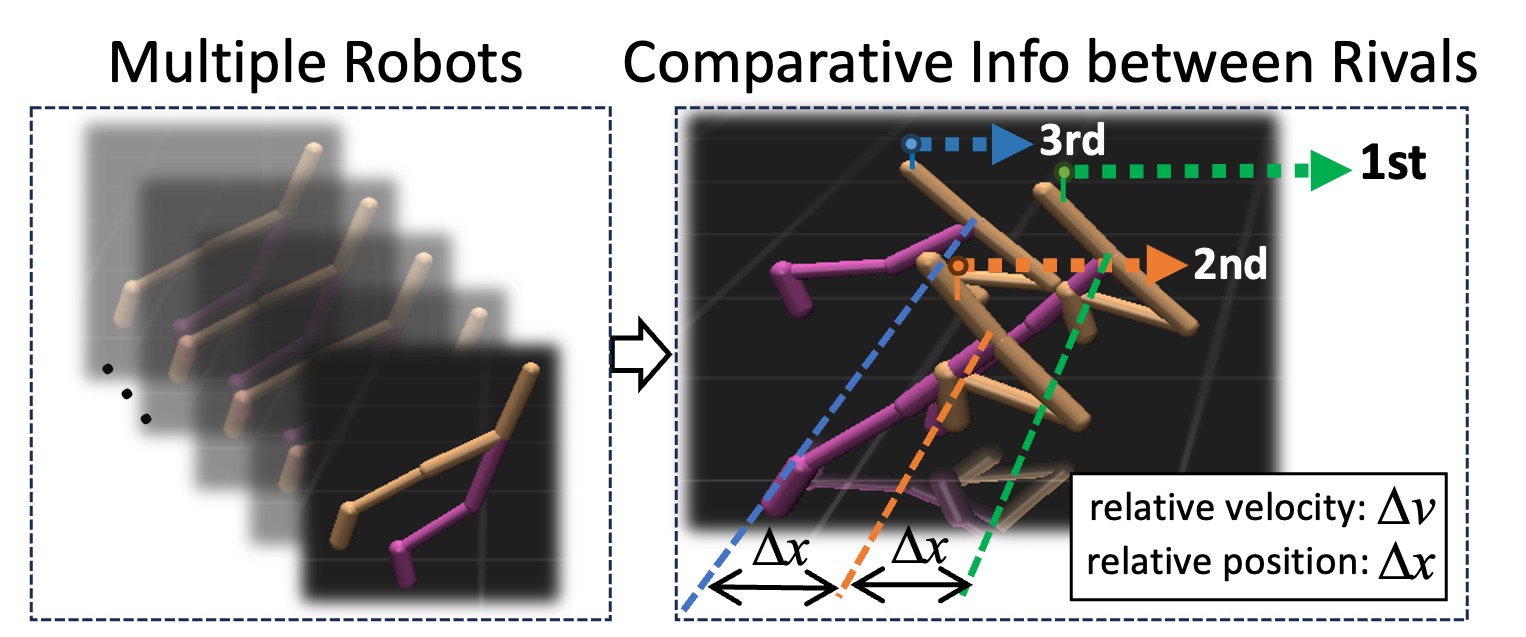}
    \caption{The proposed Self-Interest Competition Environments \textbf{Multiagent-Race}. Three walker2d robots racing is illustrated as an example.}
    \label{fig:walker2d_sim}
\end{figure}

\subsection{Environment}
We first build a series of self-interested and competitive environments named \href{https://github.com/KJaebye/Multiagent-Race}{Multiagent-Race} (\textbf{Race}). We extend the single-robot running to multi-robot racing. It is important to underline that environments are not the straightforward parallel environments of many solitary agents for sampling speed-up like what has been done in NVIDIA Isaac-Gym, but a real multiplayer competitive racing game, shown in Fig.\ref{fig:walker2d_sim}. 

We provide six \textbf{Race} environments including \texttt{MultiAnt}, \texttt{MultiWalker2d}, \texttt{MultiHalfCheetah}, \texttt{MultiHopper}, \texttt{MultiSwimmer}, \texttt{MultiHumanoid}. In \textbf{Race}, self-interested agents face the same target: reach the maximum speed within a limited duration. Relative position and velocity along the desired orientation are treated as competitive information, fed to each robot. The rewarding for each type of robot maintains alignment with the Gym. We conduct experiments on four tasks: \texttt{MultiAnt}, \texttt{MultiHopper}, \texttt{MultiWalker2d}, and \texttt{MultiCheetah}. To avoid uncertainty, we take the average of over 10 trials using random seeds.

\begin{figure*}
    \subfloat[\texttt{MultiAnt} task]{
        \includegraphics[width=4.3cm]{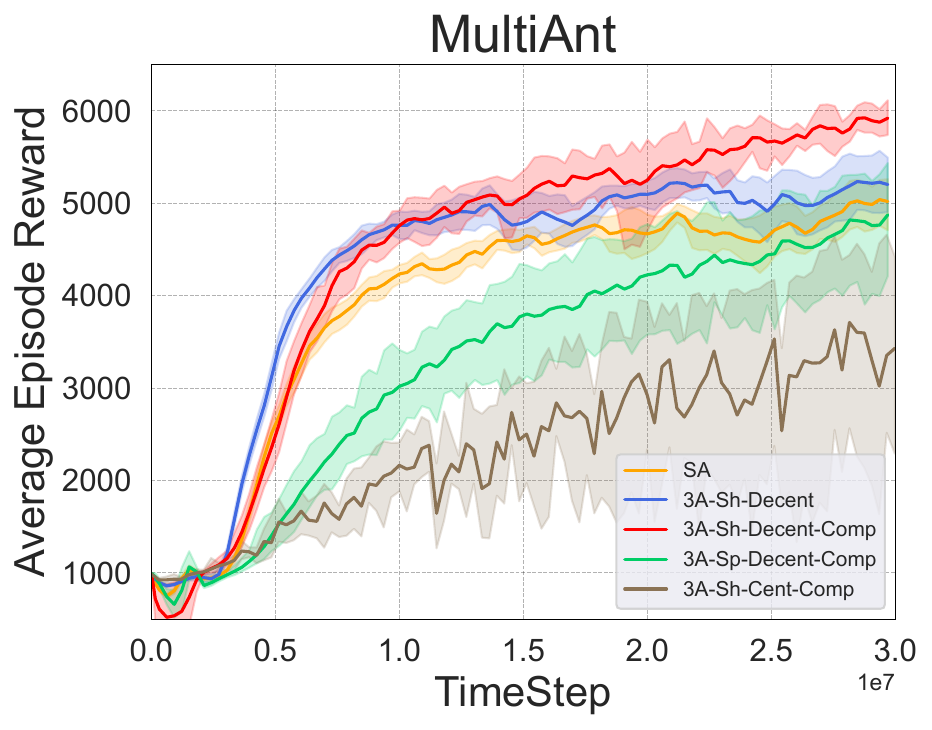}}
    \subfloat[\texttt{MultiHalfCheetah} task]{
        \includegraphics[width=4.3cm]{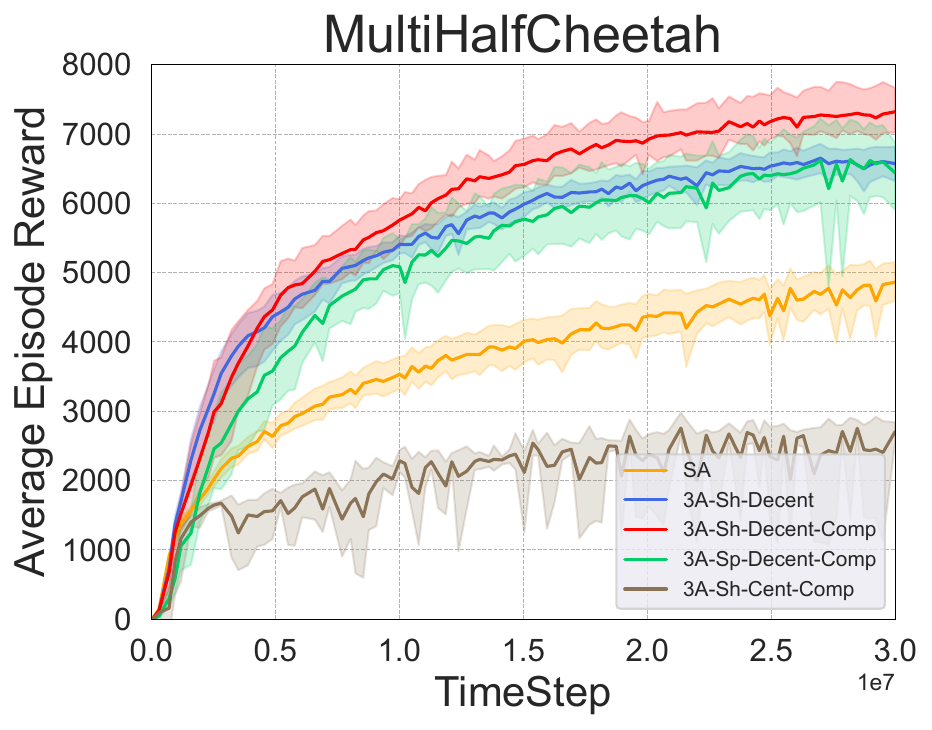}}
    \subfloat[\texttt{MultiWalker2d} task]{
        \includegraphics[width=4.3cm]{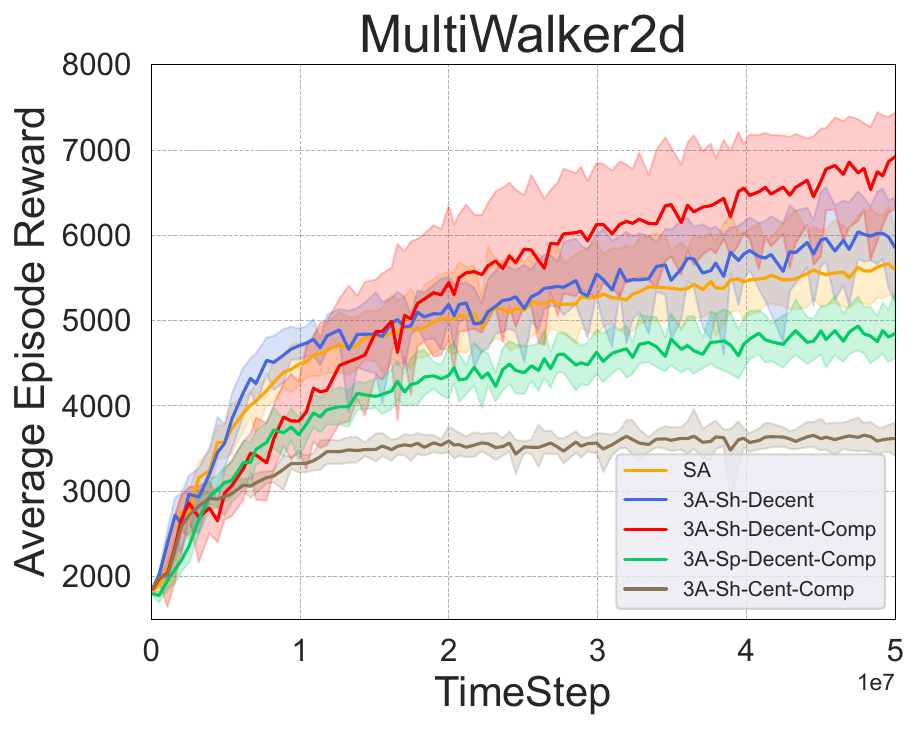}}
    \subfloat[\texttt{MultiHopper} task]{
        \includegraphics[width=4.3cm]{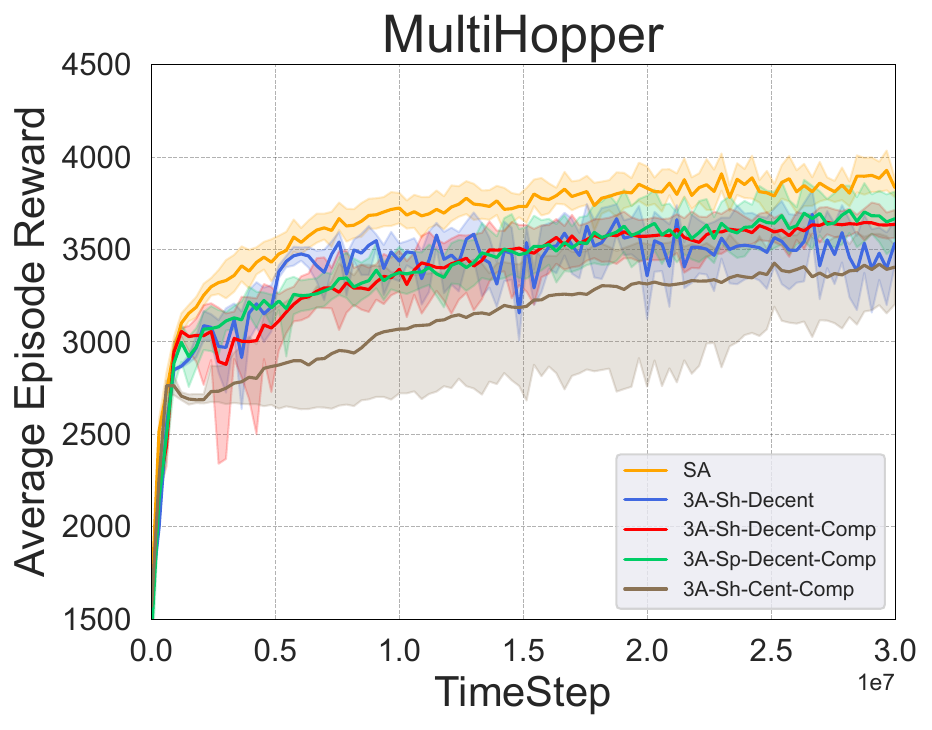}}
    \caption{Performance of different settings on 3-agent environments. $S$A: \colorbox{orange}{Yellow}. $3$A-Sh-Decent: \colorbox{royalblue}{Blue}. $3$A-Sh-Decent-Comp: \colorbox{red}{Red}. $3$A-Sp-Decent-Comp: \colorbox{green}{Green}. $3$A-Sh-Cent-Comp: \colorbox{brown}{Brown}.}
    \label{fig:main comparison}
\end{figure*}

\subsection{Baselines}
We primarily consider several benchmarks for comparison to explore the effectiveness of our approach. $\bm{\theta}$ and $\bm{\phi}$ are used to represent shared parameters, then $\theta^{i}$ and $\phi^{i}$ denote separated parameters. $X$ represents the robot number in \textbf{Race}. Our experiments are conducted in 2, 3, 4, and 5 robots race respectively, where $X=2, 3, 4, 5$, to show the effect of different scales of competition.

\textbf{$S$A:} PPO training in a single-agent environment. We have $\pi \left(a\mid s; \bm{\theta} \right)$ and $v(s;\bm{\phi})$.

\textbf{$X$A-Sh-Decent:} training in $X$ number of agents environment but with no competitive information. Here we define the knowledge-level information-sharing that agents share the same networks and experiences. The agent uses its decentralized local state as critic input. Where we have $\pi \left(a^{i}\mid s^{i}; \bm{\theta} \right)$ and $v(s^{i};\bm{\phi})$.

\textbf{$X$A-Sh-Cent:} training in $X$ number of agents environment but without competitive information. Policy and experience are shared. Agents use a centralized global state as critic input. We have $\pi \left(a^{i}\mid s^{i}; \bm{\theta} \right)$ and $v(s;\bm{\phi})$, in which $s$ is the global observation, denotes the concatenation of $(s^{1},..., s^{N})$.

\textbf{$X$A-Sp-Decent-Comp:} Training in $X$ number of agents environment with competition. Every agent trains its own specific policy using local observation as value input. We can simply consider it to be a parallel training of $S$A with competitive observation. It is different from the proposed approach: although competitive information is applied to agents, no contrastive representation is formed because experiences are not shared. Therefore, the agent cannot improve itself by learning from others. Where we have $\pi \left(a^{i}\mid \bar{s}^{i}; \theta^{i} \right)$ and $v(\bar{s}^{i};\phi^{i})$.

\textbf{$X$A-Sh-Decent-Noi:} Training in $X$ number of agents environment with random noise. Previous studies argued that noise has a significant performance improvement on multi-agent learning rather than global information\cite{Hu2021}. To validate that agents indeed acquire effective experience from competitive information, we replace with zero-mean random noise as a control group. Where we have $\pi \left(a^{i}\mid [s^{i}, n]; \bm{\theta} \right)$ and $v([s^{i}, n];\bm{\phi})$, in which $n$ denotes the noise.

\textbf{$X$A-Sh-Decent-Comp (Proposed):} Training in $X$ number of agents environment with competitive input. Agents share policy and experience.  Where we have $\pi \left(a^{i}\mid \bar{s}^{i}; \bm{\theta} \right)$ and $v(\bar{s}^{i};\bm{\phi})$.

\subsection{Experimental Details}
\textbf{Networks:}
We use the same actor-critic network structure as stable-baseline3\cite{Raffin2020} and TianShou\cite{Weng2022}: 2 hidden layers MLP with 64 units each, and Tanh activation which is then fed into the Gaussian policy action out layer (except the \texttt{MultiCheetah} task where we use Beta distribution policy).

\textbf{Algorithm Parameters:}
Routine hyperparameter settings: Adam optimizer with learning rate 0.0005 and linear learning rate decay strategy. The clipping parameter is 0.2, discounting factor is 0.995, and generalized advantage estimate parameter is 0.95. Our sampling number is larger than Tianshou and Stable-Baseline3 because we do not use the mini-batch update method for each iteration, but our optimization number is much less than theirs. However, this does not affect our comparison once they have converged.


\subsection{Role of Competition}
\begin{figure}
    \centering
    \subfloat[]{\includegraphics[height=2.3in]{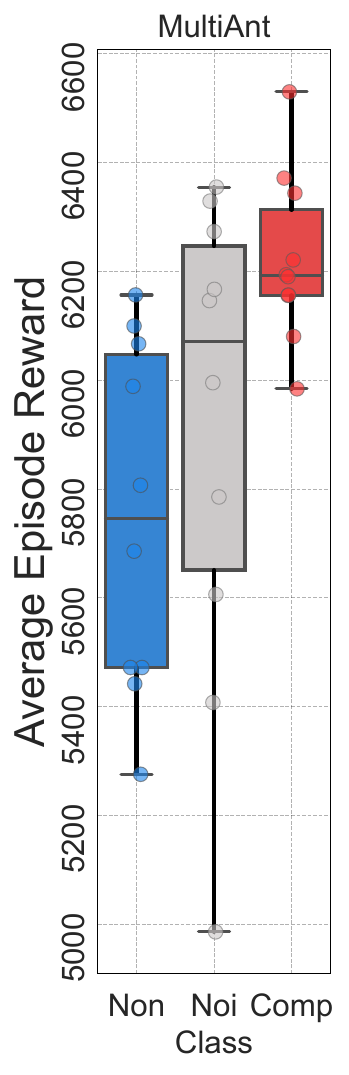}}
    \subfloat[]{\includegraphics[height=2.3in]{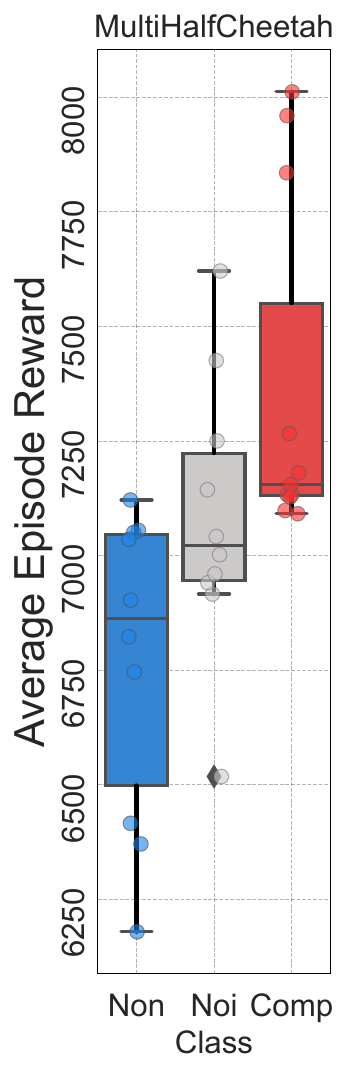}}
    \subfloat[]{\includegraphics[height=2.3in]{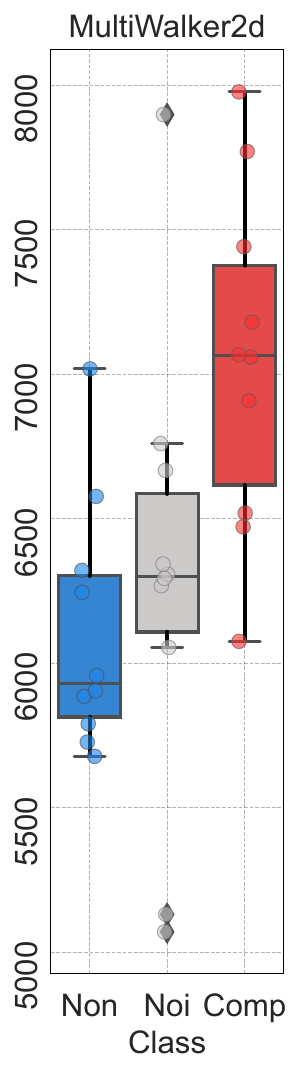}}
    \subfloat[]{\includegraphics[height=2.3in]{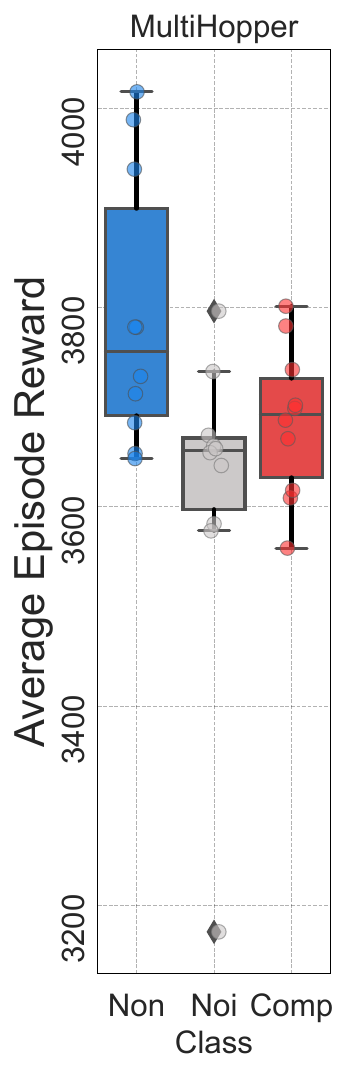}}
    \caption{Comparison between $3$A-Sh-Decent (\textbf{Non}: non-competitive information as inputs), $3$A-Sh-Decent-Noi (\textbf{Noi}: noise information as inputs), and $3$A-Sh-Decent-Comp (\textbf{Comp}: competitive information as inputs).}
    \label{fig:learn from competition}
\end{figure}

\begin{figure*}
    \subfloat[\texttt{MultiAnt} task]{
        \includegraphics[width=4.3cm]{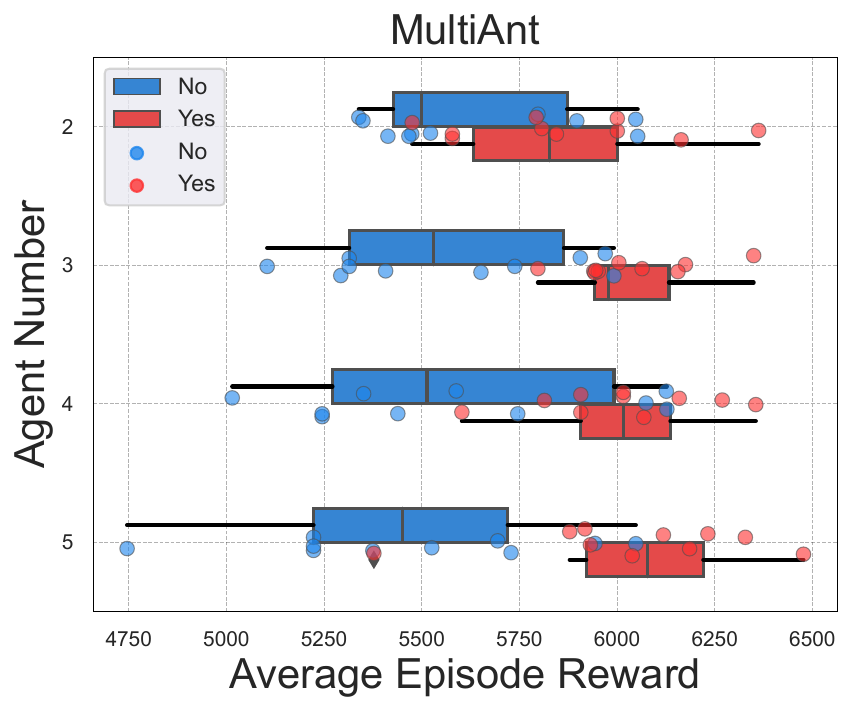}}
    \subfloat[\texttt{MultiHalfCheetah} task]{
        \includegraphics[width=4.3cm]{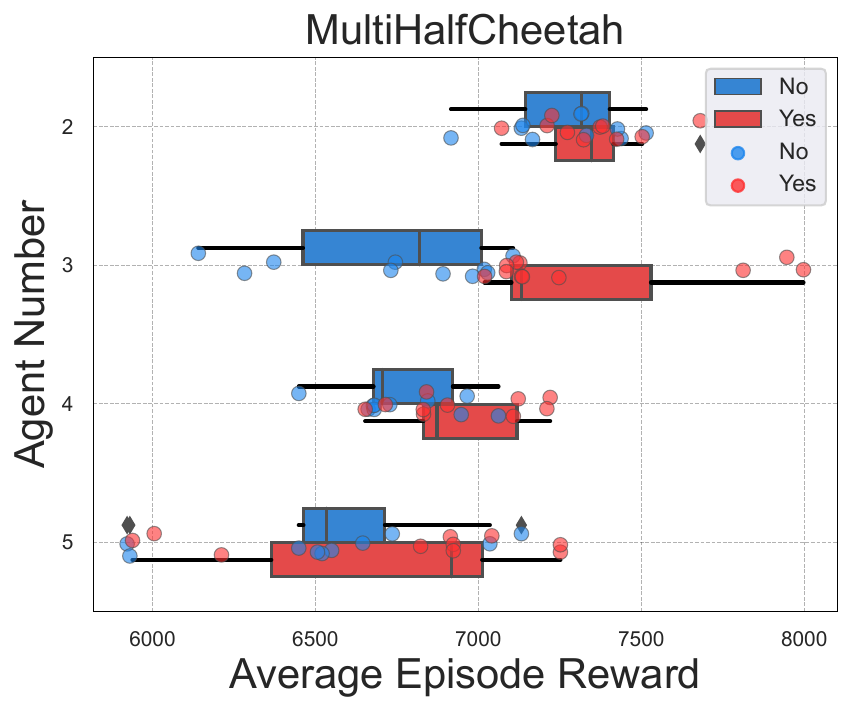}}
    \subfloat[\texttt{MultiWalker2d} task]{
        \includegraphics[width=4.3cm]{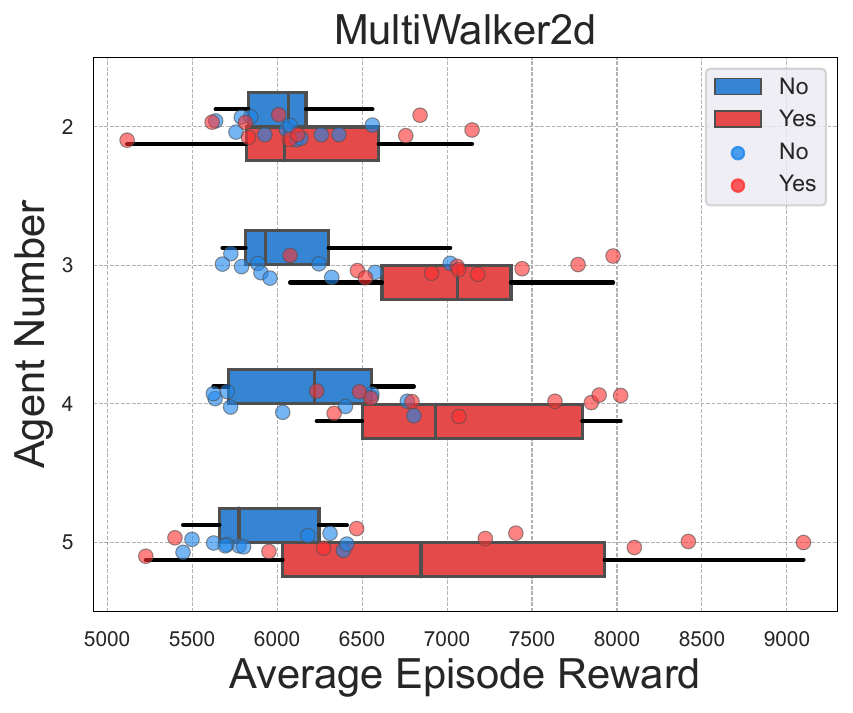}}
    \subfloat[\texttt{MultiHopper} task]{
        \includegraphics[width=4.3cm]{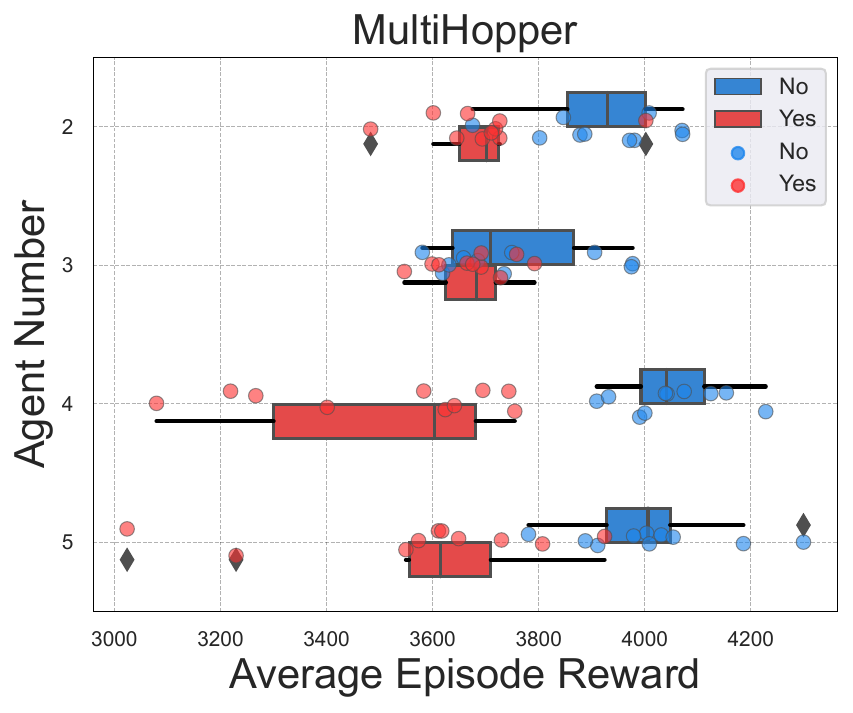}}
    \caption{Results on various numbers of agents with/without competitive info.}
    \label{fig:num compare}
\end{figure*}

We benchmark the proposed approach against baselines using 3-robot environments as a representative case. It is important to note that our method might yield better performance in environments with a different number of agents. Results are illustrated in Fig.\ref{fig:main comparison}. Our proposed approach, training with competition and shared policy ($3$A-Sh-Decent-Comp), outperforms all baselines on task \texttt{MultiAnt}, \texttt{MultiHalfCheetah}, and \texttt{MultiWalker2d}. Besides, we notice that $S$A and $3$A-Sh-Decent have a similar trend, but the latter slightly exceeds the former. This occurs because multiple agents gather more data leading to a more balanced distribution of data abundance and variance and increasing accuracy in the sampling results. Comparing $3$A-Sh-Decent and $3$A-Sh-Decent-Comp, we can conclude that robots learn valuable knowledge from additional comparative information. Especially in task \texttt{MultiAnt}, \texttt{MultiHalfCheetah}, and \texttt{MultiWalker2d}, we observe around $13\%$, $11\%$, and $16\%$ improvement respectively.

However, it does not work on \texttt{MultiHopper}. In contrast, settings of competitive training seem to negatively affect the results. This phenomenon could be attributed to the \texttt{Hopper} task being overly simplistic, and the competitive observation might drown out the proprioceptive state. On the other hand, it appears that the dynamics of the \texttt{Hopper} has been thoroughly explored and understood, leading to little improvement under competition.

\begin{table*}
\centering
\caption{Comparison with SoTA PPO Baselines. Task1: \texttt{Ant}, Task2: \texttt{Walker2d}, Task3: \texttt{HalfCheetah}, Task4: \texttt{Hopper}. A: \textbf{OpenAI}\cite{Schulman2017}, B: \textbf{Stable-Baselines3}\cite{Raffin2020}, C: \textbf{Tianshou}\cite{Weng2022}. $X$A-Sh-Decent-Comp is the proposed approach.}

\begin{tabular}
{|wc{0.02cm}|wc{0.25cm}|wc{0.9cm}|wc{1cm}|wc{0.9cm}|wc{0.9cm}|wc{0.9cm}|wc{0.9cm}|wc{0.9cm}|wc{1.1cm}|wc{1cm}|wc{0.9cm}|wc{1cm}|wc{0.9cm}|} 

\hline
\multirow{2}{*}{} & \multicolumn{3}{c|}{SOTA PPO Baselines} & \multirow{2}{*}{$S$A}                                   & \multicolumn{4}{c|}{$X$A-Sh-Decent}                                                                   & \multirow{2}{*}{\textit{3}A-Decent-Noi} & \multicolumn{4}{c|}{\textit{X}A-Sh-Decent-Comp}                                                                                                                                 \\ 
\cline{2-4}\cline{6-9}\cline{11-14}
                       & A    & B            & C                 &                                                        & 2            & 3            & 4                                                & 5            &                                 & 2                                                & 3                                                & 4             & 5                                                 \\ 
\hline
1                      & –    & 1327$\pm$452 & 3900$\pm$850      & \begin{tabular}[c]{@{}c@{}}4993$\pm$326\\\end{tabular} & 5468$\pm$371 & 5552$\pm$441 & 5987$\pm$512                                     & 5728$\pm$237 & 5781$\pm$365                    & 5926$\pm$456                                     & 5960$\pm$207                                     & 6056$\pm$382  & {\cellcolor[rgb]{0.275,0.941,0.275}}\textbf{6140$\pm$302}  \\ 
\hline
2                      & 3424 & 3479$\pm$822 & 4896$\pm$704      & 5579$\pm$488                                           & 6107$\pm$463 & 5902$\pm$665 & 6491$\pm$637                                     & 5711$\pm$441 & 6311$\pm$412                    & 5876$\pm$1019                                    & {\cellcolor[rgb]{0.275,0.941,0.275}}\textbf{7094$\pm$945} & 6683$\pm$1379 & 6634$\pm$2039                                     \\ 
\hline
3                      & 1669 & 5819$\pm$663 & 7337$\pm$1508     & 4988$\pm$334                                           & 7284$\pm$289 & 6706$\pm$519 & 6671$\pm$293                                     & 6505$\pm$569 & 7063$\pm$363                    & {\cellcolor[rgb]{0.275,0.941,0.275}}\textbf{7342$\pm$213} & 7197$\pm$498                                     & 6790$\pm$277  & 6873$\pm$702                                      \\ 
\hline
4                      & 2316 & 2410$\pm$10  & 3128$\pm$413      & 3834$\pm$313                                           & 3913$\pm$187 & 3701$\pm$201 & {\cellcolor[rgb]{0.275,0.941,0.275}}\textbf{4059$\pm$165} & 4002$\pm$205 & 3679$\pm$89                     & 3715$\pm$54                                      & 3688$\pm$173                                     & 3598$\pm$267  & 3606$\pm$77                                       \\
\hline

\end{tabular}
\label{table:compare}
\end{table*}
\subsection{Ablation Studies}
We perform a series of ablation studies to obtain a more profound understanding of how competitive scenarios yield valuable data for the learning process.

\textbf{Policy and Buffer Sharing:} We conducted controlled experiments on four environments depending on whether the policy network and experience reply buffer are shared or not, using the 3-robot \textbf{Race} game as an example. In $3$A-Sp-Decent-Comp, robots enjoy independent policies, while robots share one policy and buffer in $3$A-Sh-Decent-Comp. Results are shown in Fig.\ref{fig:main comparison}. Irrespective of the varying tasks, agents with a shared network consistently demonstrate superior performance compared to their counterparts utilizing independent policy. This substantiates the explanation of (\ref{eq:critic}) discussed above.

Moreover, as there is no interaction and cooperation between competitors, the value network need not coordinate the centralized global states of all robots. By contrast, centralized value inputs might confuse agents and harm the performance of evaluation, illustrated as brown lines in Fig.\ref{fig:main comparison}.

\textbf{Learning from Competition:} 
There remains ongoing debate regarding the influence of external information on the effectiveness of training. Some studies have proved that adding more comprehensive information to the MARL task as additional states could improve the performance\cite{DeWitt2020, Yu2021}. However, others believe the network might treat external messages as noise that encourages more exploration\cite{Hu2021}, rather than learning valuable features from messages.

We employ random noise that is equivalent in length to the competitive information message as a controlled group, named $X$A-Sh-Decent-Noi. $3$S-Sh-Decent is also taken as the comparison using fixed zero padding instead. We compare the performance of $3$A-Sh-Decent, $3$A-Sh-Decent-Noi, and $3$A-Sh-Decent-Comp in Fig.\ref{fig:learn from competition}. Our results can corroborate the viewpoints presented in \cite{Hu2021}: assigning noisy observations as inputs enhances the network's exploration capability, consequently leading to higher rewards. Simultaneously, we also prove that agents can acquire knowledge from competitive observation, leading to a higher reward surpassing those of the noise experiments.

\section{Discussion}
We find there exists an unavoidable correlation between the results of games, the complexity of robots, and the number of racers involved. To explore how the number of racers affects the results, we conduct experiments with different numbers of participants, shown in Fig.\ref{fig:num compare}.

We compare the outcomes of utilizing competitive information in different environments and varying numbers of runners in \textbf{Race} tasks. For the sake of fairness, we record data after convergence. The results shown in Fig.\ref{fig:num compare} indicate that training with competitive pressure can promote performance regardless of racer number, except \texttt{Hopper} because we believe its potential has been fully explored. Besides, in Table.\ref{table:compare}, we also compare the experimental results with the state-of-the-art PPO baselines. Our proposed method makes a great improvement on \texttt{Walker2d} and \texttt{Ant}, then a slight improvement on \texttt{HalfCheetch} task.

Furthermore, we discovered that on more challenging tasks like more complex robots, the advantages of utilizing competition are more pronounced. For example, robot \texttt{Ant} simulates in the three-dimensional environment which is more complex, robots achieve higher running speeds with an increase in the number of competitors. However, for agents with simpler structures, indiscriminately increasing the number of competitors could potentially lead to a decrease on final rewards. Declines are observed on 2D robots \texttt{Hopper}, \texttt{HalfCheetah}, and \texttt{Walker2d}. This is because robots with simple structures and straightforward dynamics can easily reach their ability boundary, while robots with complex structures and challenging tasks require more sophisticated controllers. We obtain the best performance on 2-\texttt{HalfCheetah}, 3-\texttt{Walker2d}, 5-\texttt{Ant}, and single-\texttt{Hopper}.

In addition, an excessive number of competitors, however, introduce an excessive amount of competitive information, resulting in an increase in the observation dimension. The additional signals representing competition could overshadow the proprioceptive signals of the agent. This leads to challenges in robot learning when the number of runners becomes excessive.

Although our work primarily focuses on PPO as the core algorithm, our framework is adaptable to any other on-policy multi-agent algorithms. On-policy algorithms, by not relying on past experiences, can establish precise comparative features within a single batch of data sampled by the current policy. The framework cannot adapt to off-policy algorithms. The preservation of old experiences is common practice in off-policy methods, leading to the absence of a baseline value for evaluating the comparison of information along a long-term training process.

\section{Conclusion}
\label{sec:conclusion}
In this work, we propose a competitive learning framework that can stimulate robots' potential. Our method effectively leverages the competition, allowing for increased exploration and exploitation of comparative data, even using raw data as additional input. Through extensive experimentation, we have empirically demonstrated that, under competitive learning among multiple self-interested racers, our method can surpass the majority of SoTA benchmarks, including Tianshou and Stable-Baselines3. In the future, how proprioceptive signal to additional signal ratio influences the training can be further explored. Besides, more experiments can be implemented to verify the effectiveness in the real world.





\bibliographystyle{IEEEtran}
\bibliography{main}  

\addtolength{\textheight}{-12cm}   

\end{document}